\documentclass{article}
\usepackage{ijcai17}

\usepackage{times}
\usepackage{epsfig}
\usepackage{graphicx}
\usepackage{amsmath}
\usepackage{amssymb}
\usepackage{mathsymb}
\usepackage{textcomp}
\usepackage{verbatim}
\usepackage{subfigure}
\usepackage{url}
\usepackage{graphicx}
\usepackage{multirow}
\usepackage[super]{nth}
\usepackage{balance}
\usepackage{colortbl}
\graphicspath{{./}{./Fig/}{./Figs/}{./Fig/eps/}{./Fig/pdf/}}
\usepackage[ruled,linesnumbered,vlined]{algorithm2e}

\newcommand{\eg}{\emph{e.g.,\ }}

\newtheorem{remark}{Remark}

\title{Beyond Low-Rank Representations: Orthogonal Clustering Basis Reconstruction with Optimized Graph Structure for Multi-view Spectral Clustering }
\author{Yang Wang$^{\dag}$,  Lin Wu$^{\ddag}$\\
 $^{\dag}$The University of New South Wales, Kensington, Sydney, Australia\\
 $^{\ddag}$The University of Queensland, Australia\\
 wangy@cse.unsw.edu.au; lin.wu@uq.edu.au
}

\begin{document}

\maketitle

\begin{abstract}
 Low-Rank Representation (LRR) is arguably one of the most powerful paradigms for Multi-view spectral clustering, which elegantly encodes the multi-view local graph/manifold structures into an intrinsic low-rank self-expressive data similarity embedded in high-dimensional space, to yield a better graph partition than their single-view counterparts.  In this paper we revisit it with a fundamentally different perspective by discovering LRR as essentially a latent clustered orthogonal projection based representation winged with an optimized local graph structure for spectral clustering; each column of the representation is fundamentally a cluster basis orthogonal to others to indicate its members, which intuitively projects the view-specific feature representation to be the one spanned by all orthogonal basis to characterize the cluster structures. Upon this finding, we propose our technique with the followings: (1) We decompose LRR into latent clustered orthogonal representation via low-rank matrix factorization, to encode the more flexible cluster structures than LRR over primal data objects; (2) We convert the problem of LRR into that of simultaneously learning orthogonal clustered representation and optimized local graph structure for each view; (3) The learned orthogonal clustered representations and local graph structures enjoy the same magnitude for multi-view, so that the ideal multi-view consensus can be readily achieved. The experiments over multi-view datasets validate its superiority,
especially over recent state-of-the-art LRR models.
\end{abstract}

\section{Introduction}
Spectral clustering \cite{nips01}, which partitions the data objects via their local graph/manifold structure relying on the Laplacian eigenvalue-eigenvector decomposition,  is one fundamental clustering problem. Unlike K-Means clustering \cite{XINDONG}, the data objects within the same group characterize not only the large data similarity but also the similar local graph/manifold structure. With the rapid development of information technology, the data are largely available with the multi-view feature representations (\eg images can be featured by a color histogram view or a texture view), which naturally paves the way to multi-view spectral clustering. As extensively claimed by the multi-view research \cite{TaoPAMIA,WangTNNLS,weiijcai15,ACMMM13,ACMSIGIR15,CIKM13,ACMMM15,PAKDD14}, the information encoded by multi-view features describe different properties; thus leveraging the multi-view information can outperform the single-view counterparts. One critical issue on a successful multi-view incorporation implied by the existing work \cite{TaoTIP14,icml11,LinIVC2017,YangTIP2015,YangTIP2017}, lies in how to achieve the multi-view consensus/agreement.

Following such principle, a lot of multi-view clustering methods \cite{SDM13,ICCVsub15} claim that similar data objects should be within the same group across all views. Based on that, the consensus multi-view local manifold structure is further explored with great efforts \cite{RMVSC,ijcai16,NIPS11,icml11}
for multi-view spectral clustering.  Among all these methods,  Low-Rank Representation (LRR) \cite{LRRICML2010} coupled with sparse decomposition based model has been emerged as a substantially elegant solution, due to its strength of exploring their intrinsic low-dimensional manifold structure encoded by the data correlations embedded in high-dimensional space, while exhibiting strong robustness to feature noise corruptions addressed by sparse noise modeling,  hence attracting great attention.
\subsection{Motivation: LRR Revisited for Multi-View Spectral Clustering}\label{sec:revisit}
Specifically, the typical LRR model for multi-view spectral clustering stems from the formulation below:
\begin{equation}\label{eq:motivate}
\begin{aligned}
& \min_{Z, E_i} ||Z||_* + \lambda \sum_{i \in V} ||E_i||_1 \\
& \textmd{s.t.}~~ X_i = X_iZ + E_i, ~~i \in V, ~~Z \succeq 0,
\end{aligned}
\end{equation}
where $X_i \in \mathbb{R}^{d_i \times n}$ is the data representation for the $i^{th}$ view with $d_i$ as its feature dimension, $n$ as the number of data objects identical for each view, $\lambda$ is the balance parameter, and $V$ is the view set. $Z \in \mathbb{R}^{n \times n}$ is the self-expressive low-rank similarity representation shared by all $|V|$ views, constrained with $||Z||_*$ based on $X_i (i \in V)$ , which can also be substituted by the other specific dictionaries; $||E_i||_1$ is modeled to address the noise-corruption for the $i^{th}$ view-specific feature representation. $Z \succeq 0$ ensures the nonnegativity for all its entries. Based on such optimized low-rank $Z$, the spectral clustering is finally conducted. One significant limitation of Eq.\eqref{eq:motivate} pointed out by \cite{ijcai16} is that, \emph{only one common} $Z$ is learned to \emph{preserve the flexible local manifold structures for all views}, hence fails to achieve the ideal spectral clustering result.
To this end, various low-rank $Z_i$ are learned to preserve the $i^{th}$ view-specific local manifold structures, meanwhile minimize their divergence via an iterative-views-agreement strategy for multi-view consensus, followed by a final spectral clustering stage.
Despite its encouraging performance, the following standout limitations are inattentively overlooked for LRR model: \textbf{(1)} The low-rank data similarity may not well encode the flexible latent cluster structures over primal view-specific feature space; worse still for the non-ideal local graph construction over such representation for spectral clustering; \textbf{(2)} The low-rank data similarities coming from multi-views may not be within the same magnitude, so that the divergence minimization may not achieve the ideal multi-view clustering consensus.\\
\textbf{\underline{Our new perspective}.} The above facts motivate us to revisit the low-rank representation $Z_i$ to help $X_iZ_i$ reconstruct $X_i$ below for the $i^{th}$ view
\begin{equation}\label{eq:trans}
\min_{Z_i \in \mathbb{S}}||X_i - X_iZ_i||_F^2,
\end{equation}
where $\mathbb{S}$ denotes the set  of $Z_i \in \mathbb{R}^{n \times n}$ with low-rankness \eg cluster number $c$ far less than $d_i$; \textbf{Instead of narrowing Low-Rank $Z_i$ as self-expressive data similarity from the conventional viewpoint, it is essentially seen as a special case of a generalized Low-Rank projection, to map feature representation to a low-dimensional space to reconstruct $X_i$ with minimum error.} As discussed, the self-expressive similarity projection equipped with LRR models still suffer from the aforementioned non-trivial limitations.

\textbf{Here we ask a question: Is there a superior low-rank projection $Z_i$ to minimize Eq.\eqref{eq:trans}, meanwhile address the limitations over the existing LRR models.} Our answer to this question is positive. Specifically, we propose to consider $Z_i$ as a latent clustered orthogonal projection, via $Z_i = U_iU_i^T$, where
\begin{enumerate}
\item \textbf{\underline{Clustered orthogonal projection}}: $U_i \in \mathbb{R}^{n \times c}$, where each column indicates one cluster to characterize its belonging data objects. Compared with LRR over original feature space, the latent factor $U_i$ can better preserve the flexible latent cluster structure.
\item \textbf{\underline{Feature reconstruction with cluster basis}}: Instead of low-rank data similarity, $Z_i$ essentially serves as a mapping to reconstruct the view-specific features via the column of $U_i$ to encode the latent cluster structures.
\item \textbf{\underline{Rethinking $X_iZ_i$}}:  We revisit the intuition of $X_iZ_i$ via $(X_iU_i)U_i^T$ throughout two stages, remind that $X_i \in \mathbb{R}^{d_i \times n}$ where
    \begin{itemize}
    \item $X_iU_i$ is performed to obtain the new projection value for all $d_i$ features over $c$ orthogonal columns of $U_i$;
    \item $X_iU_iU_i^T$ is subsequently the projected representation for all $d_i$ features spanned by $c$ clustered orthogonal column basis of $U_i$.
    \end{itemize}
\item \textbf{\underline{Same magnitude for multi-view consensus}}: All $U_i(i \in V)$ enjoy the same magnitude due to their orthonormal columns. Hence, the feasible divergence minimization will facilitate the multi-view consensus.
\end{enumerate}
Before shedding light on our technique, we review the typical related work for multi-view spectral clustering
\subsection{Prior Arts}\label{sec:related}
The prior arts can be classified as per the strategy at which the multi-view fusion takes place for spectral clustering.

The most straightforward method goes to the \emph{Early fusion} \cite{cvpr10} by concatenating the multi-view feature vectors with equal or varied weights into an unified one, followed by the spectral clustering over such unified space. However, such method ignores the statistical property belonging to an individual view. \emph{Late fusion} \cite{ecmlpkdd09} may address the limitation to some extents by  aggregating the spectral clustering result from each individual view, which follows the assumption that all views are independent to each other. Such assumption is not effective for multi-view spectral clustering as they assume the views to be dependent so that the multi-view consensus information can be exploited for promising performance.

\emph{Canonical Correlation Analysis} (CCA) is applied for multi-view spectral clustering \cite{ICML09} by learning a common low-dimensional representations for all views, upon which the spectral clustering is performed. One salient drawback lies in the failure of preserving the flexible local manifold structures for different views via such common subspace. \emph{Co-training} based model \cite{icml11} learned the Laplacian eigenmap for each view over its projected data representation throughout the laplacian eigenmaps from other views, such process repeated till the convergence, the final similarity are then aggregated for spectral clustering. A similar method \cite{NIPS11} is also proposed to coordinate multi-view laplacian eigenmaps consensus for spectral clustering. Despite their effectiveness, they have to follow the scenario of noise free for the feature representations. Unfortunately, it cannot be met in practice.  The Low-Rank Representation and sparse decomposition models \cite{ijcai16,RMVSC} well tackle the problem, meanwhile exhibits the robustness to feature noise corruptions. However, they still suffer from the aforementioned limitations. To this end, we make the following orthogonal contributions to typical LRR model for multi-view spectral clustering.
\vspace{-0.1cm}
\subsection{Our Contributions}
\begin{itemize}
\item We revisit the classical Low-Rank Representation (LRR) for multi-view spectral clustering with a fundamentally novel viewpoint of finding it as essentially the latent clustered orthogonal projection based representation with optimized graph structure, to better encode the flexible latent cluster structures than LRR over primal data objects.
\item We convert the problem of learning LRR into that of simultaneously learning the clustered orthogonal representation and its optimized local graph structure for each view, rather than directly rely on the local graph construction over original data objects.
\item The learned multi-view latent clustered representations and local graph structures enjoy the same magnitude, so as to facilitate a feasible divergence minimization to achieve superior multi-view consensus for spectral clustering.
\end{itemize}
Extensive experiments over multi-view datasets validate the superiority of our method.

\section{Learning Clustered Orthogonal projection with Optimized Graph Structure}
In this section, we formally discuss our technique. Some notations that are used throughout the paper are shown below.
\subsection{Notations}
For Matrix $M$, the trace of $M$ is denoted as $Tr(M)$;$||M||_F = \sqrt{\sum_{i,j}M_{i,j}^2})$
(or $||\cdot||_2$ for vector space) denotes the Frobenius norm;
$||M||_1(\sum_{i,j}|M_{i,j}|)$ is the $\ell_1$ norm, and $M^T$ denotes the transpose of $M$,
and its unclear norm as $||M||_*$ (sum of all singular values); $M(i, \cdot)$ and $M(\cdot, i)$ as the $i^{th}$ row and column of $M$. $M \succeq 0$ means all entries of $M$ are nonnegative. $I$ is the identity matrix with adaptive size. \textmd{\textbf{1}} indicates the vector of adaptive length with all entries to be 1. $|\cdot|$ indicates the cardinality of the set.

\subsection{Problem Formulation}
As previously defined in section ~\ref{sec:revisit}, $X_i$ is the data representation for the $i^{th}$ view. $Z_i \in \mathbb{R}^{n \times n}$ is low-rank data similarity representation for $X_i$.
The Eq.\eqref{eq:trans} is equivalent to computing $U_i \in \mathbb{R}^{n \times c}$ such that  $Z_i = U_iU_i^T$, where $U_i$ has orthonormal columns with its $g^{th}$ column representing the relevance each data object belongs to the $g^{th}$ cluster, and $c$ indicates latent cluster number. We then arrive at the following
\begin{equation}\label{eq:reconstruct}
\min_{U_i}||X_i - X_iU_iU_i^T||_F^2
\end{equation}
As discussed in section \ref{sec:revisit}, $X_iU_iU_i^T$ reveals the new projection representation for all $d_i$ features spanned by the orthogonal basis of $U_i$ to reconstruct $X_i$.

Optimizing Eq.\eqref{eq:reconstruct} w.r.t. $U_i$ is equivalent to computing an SVD of $X_i$ to constitute the orthogonal columns of $U_i$ using the principle eigenvectors.  Inspired by this, we exploit the latent cluster structures of $X_i$ to form non-overlapping clusters with each characterized by one orthogonal column basis of $U_i$. Thanks to \cite{nuclearfactorize} on low-rank matrix factorization, it yields the following
\begin{equation}
||Z_i||_* = \min_{U_i, V_i, Z_i = U_iV_i^T} \frac{1}{2}(||U_i||_F^2 + ||V_i||_F^2),
\end{equation}
where $U_i \in \mathbb{R}^{n \times c}$ and $V_i \in \mathbb{R}^{n \times c}$ are latent factors from $Z_i$.  Based on that, we approximate $||Z_i||_*$ via the clustered orthogonal projection factorization $Z_i = U_iU_i^T$, and convert the problem of minimizing $||Z_i||_*$ to that of learning clustered projection representation $U_i$ below
\begin{equation}\label{eq:Ui}
||Z_i||_* = \min_{U_i, Z_i = U_iU_i^T} ||U_i||_F^2
\end{equation}
\vspace{-0.4cm}
\begin{remark}
Unlike the data similarity over raw data objects, $U_i$ via the low-rank matrix factorization can achieve the flexible latent cluster structures. Another crucial issue left to be addressed lies in its local manifold/graph structure modeling over $U_i$, which is crucial for spectral clustering. \textbf{One may directly refer to the local graph construction over $X_i$. However, as previously stated, it cannot effectively encode the the local graph structure over $U_i$.}
\end{remark}
Towards this end, we propose to learn an optimized local graph structure over $U_i$ by solving the following
\begin{align}\label{eq:localstructure}
& \frac{1}{2}\min_{\forall j \sum_{k}^{n}W_i(j,k) = 1, W_i \succeq 0}\sum_{j,k}^{n}||U_i(j,\cdot) - U_i(k,\cdot)||_2^2W_i(j,k) \\ \nonumber
& ~~~~~~= Tr(U_i^TL_iU_i),
\end{align}
where $L_i = D_i - \frac{W_i^T + W_i}{2}$ is Laplacian matrix, $D_i$ is the diagonal matrix with its $g^{th}$ diagonal entry equaled to the sum of the $g^{th}$ row of $\frac{W_i^T + W_i}{2}$. The ideal $W_i$ reveals the probability of $j^{th}$ and $k^{th}$ data points within the same cluster according to cluster projection representation $U_i$.
We impose the constraint that $\forall j$, $\sum_{k}^{n}W_i(j,k) = 1$ and $W_i(j,\cdot) \succeq 0$ to meet the probability nature of $W_i$. Following \cite{KDD14}, we will impose the regularization $||W_i||_F^2$ to avoid that only the nearest neighbor of each data point is assigned 1 with others 0.

With all the above collected, we finally formulate the problem below
\begin{small}
\begin{align}\label{eq:finalobj}
& \min_{U_i, E_i, W_i (i \in V)} \sum_{i \in V}(\underbrace{||E_i||_1}_\text{sparse noise modeling}  + \underbrace{\lambda_1||W_i||_F^2}_\text{regularized graph structure}\\ \nonumber
& + \underbrace{\frac{\lambda_2}{2}Tr(U_i^TL_iU_i)}_\text{structuring $U_i$ with optimized local manifold structure}   \\ \nonumber
& + \underbrace{\frac{\beta}{2}\sum_{j \in V, j \neq i}||U_i - U_j||_F^2}_\text{modeling the multi-view consensus over the $U_i$ within the same magnitude}) \\ \nonumber
& ~~~~\textmd{s.t.}~~~~ i = 1, \ldots, V, ~~~X_i = D_iU_i^T + E_i, D_i = X_iU_i\\ \nonumber
& ~~U_i^TU_i = I, G_i = U_i, G_i \geq 0, \forall j \sum_{k = 1}^{n}W_i(j,k) = 1; W_i \succeq 0,\\ \nonumber
\end{align}
\end{small}
where $||U_i||_F^2$ is omitted due to constraint $U_i^TU_i = I$; all the $U_i \in \mathbb{R}^{n \times c} (i \in V)$ share the same cluster number $c$ for multi-view clustering consensus. $\lambda_1$, $\lambda_2$ and $\beta$ are non-negative weights related to learning the clustered orthogonal representation, its local graph structure and multi-view consensus modeling, and will be studied in Section \ref{sec:experiment}. The constraint $U_i^TU_i = I$ ensures the orthonormal columns of $U_i$.
\vspace{-0.1cm}
\begin{remark}
We introduce two auxiliary variables $G_i = U_i$ and $D_i = X_iU_i \in \mathbb{R}^{d_i \times c}$. As will be shown later, the intuition of introducing $D_i$ lies in minimizing  $||X_i - D_iU_i^T  - E_i||_F^2$ w.r.t. $D_i$, where
\begin{itemize}
\item it is similar as dictionary learning, while popping up $U_i^T$ as the corresponding sparse representation learning; moreover, it also enjoys the optimization of the isolated $U_i^T$ after merging the other $U_i$ into $D_i$.
\end{itemize}
\end{remark}
\section{Optimization}
Solving Eq.\eqref{eq:finalobj} is equivalent to be a unified process of simultaneously learning $U_i$ and $W_i$ for the $i^{th} (i \in V)$ view. As will be shown later, learning either of them will promote the other.  Optimizing Eq.\eqref{eq:finalobj} is not jointly convex to $U_i$, $W_i$ and $E_i$, we hence alternately optimize each of them with the others fixed. Following \cite{ALM11}, we deploy the Augmented Lagrange Multiplier (ALM) together with Alternating Direction Minimization (ADM) strategy, which is widely known as an effective and efficient solver. As the optimization process for the above variables within each view is similar, we only present the optimization process for the $i^{th}$ view, the same process holds for other $j^{th}(j \neq i)$views.  The augmented lagrangian function can be written below
\begin{small}
\begin{align}\label{eq:lagran}
& \min_{ \forall j \sum_{k = 1}^{n}W_i(j,k) = 1, 0 \leq W_i(j,k) \leq 1, U_i^TU_i = I}\mathcal{L}(U_i, E_i, D_i, G_i, W_i) \\ \nonumber
& = ||E_i||_1 + \frac{\lambda_2}{2}Tr(U_i^TL_iU_i) + \lambda_1||W_i||_F^2\\ \nonumber
& + \frac{\beta}{2}\sum_{i \in V, i \neq j}||U_i - U_j||_F^2 + \Phi(K_1^i, X_i - D_iU_i^T - E_i) \\ \nonumber
& + \Phi(K_2^i, U_i - G_i) + \Phi(K_3^i, D_i - X_iU_i) \\ \nonumber
& + \Phi(\mu, ||X_i - D_iU_i - E_i||_F^2 + ||U_i - G_i||_F^2 + ||D_i - X_iU_i||_F^2), \\ \nonumber
\end{align}
\end{small}
where $K_1^i \in \mathbb{R}^{d_i \times n}$, $K_2^i \in \mathbb{R}^{n \times c}$, and $K_3^i \in \mathbb{R}^{d_i \times c}$ are Lagrange multipliers. $\Phi(\cdot,\cdot)$ indicates element-wise multiplication. $\mu > 0$ is a penalty parameter.\\
\underline{\textbf{Solving $U_i$}}: We calculate the partial derivative of Eq.\eqref{eq:lagran} w.r.t. $U_i$, $\frac{\partial{\mathcal{L}}}{\partial{U_i}}$ to be $\textbf{0} \in \mathbb{R}^{n \times c}$, while fixing others to be constant. After rearranging the terms, it has
\begin{equation}\label{eq:Uideri}
U_i = (\underbrace{\lambda_2L_i + (\mu + \beta(|V| - 1))I - \mu X_i^TX_i)^{-1}}_\text{With $\mathcal{O}(n^3)$ computational complexity}S,
\end{equation}
where
\begin{align}\label{eq:SS}\nonumber
\tiny
& S = \sum_{j \in V, j \neq i} U_j + ((K_1^i)^T - \mu U_iD_i^T - \mu E_i^T)D_i \\ \nonumber
& ~~+ X_i^TK_3^i + \mu X_i^TX_iU_i \\ \nonumber
\end{align}
\textbf {\underline{Efficient Row updating strategy of $U_i$}}.
As shown in Eq.\eqref{eq:Uideri}, the bottleneck of updating $U_i$ lies in the high computational complexity of $\mathcal{O}(n^3)$ caused by the matrix inverse operation against the $\mathbb{R}^{n \times n}$. To resolve it, we propose to update each row of $U_i$. Without loss of generality, we set the derivative w.r.t. $U_i(l,\cdot), 1 \leq l \leq n$ to be $\textbf{0} \in \mathbb{R}^{c}$. It then yields the following
\begin{equation}\label{eq:col}
U_i(l,\cdot)
= (T_i^l  + \beta\underbrace{\sum_{j \neq i, j \in V}U_j(l,\cdot)}_\text{Influences from other views})
\underbrace{(R_i + D_i^TD_i)^{-1},}_\text{computational complexity $\mathcal{O}(c^3)$}
\end{equation}
where
\begin{equation}\label{eq:Ri}
R_i = ((1 + \mu + \sum_{k = 1}^{n}(\lambda_2L_i(k,l) - \mu (X_i^TX_i)(k,l)))I \in \mathbb{R}^{c \times c}
\end{equation}
\begin{align}\nonumber
& T_i^l = X_i^T(l,\cdot)K_3^i + \mu \left(G_i(l,\cdot) - E_i^T(l,\cdot)D_i\right)\\ \nonumber
& ~~~- K_2^i(l,\cdot) - (K_1^i)^T(l,\cdot)D_i \nonumber
\end{align}
\underline{\textbf{Orthonormalize $U_i$}}: After obtaining the whole $U_i$ by updating all rows for each iteration,
the clustering algorithm \eg fast k-means is performed, which yields the cluster indicator for each data point/each row, leading to orthogonal columns
then normalize each entry of $U_i$ via the rules as: $U_i(j,k) = \frac{1}{\sqrt{|C_k|}}$ if $x_j$ is assigned with the $k^{th}$ cluster $C_k$, it is 0 otherwise. According to the processing above, it successfully achieves the orthonormal columns of $U_i$ ($i = 1, \ldots, |V|$). \\

\begin{remark}\label{re:con}
As per the row-update strategy for $U_i$ in Eq.\eqref{eq:col}, we remark the followings:
\begin{enumerate}
\item We dramatically reduces the computational complexity from $\mathcal{O}(n^3)$ by Eq.\eqref{eq:Uideri} to $\mathcal{O}(c^3)$, due to $c \ll n$.
\item Another note goes to the process of multi-view consensus of $U_i$ via the row update.
Specifically, during each iteration, the $U_i(l,\cdot)$ is updated via the influence from other views, while served as a constraint to guide the $U_j(l,\cdot) (j \neq i)$ updating, among all of which the divergence is decreased towards a consensus, \textbf{which is based on the same magnitude among $U_i(i \in V)$ with orthonormal columns.}
\end{enumerate}
\end{remark}
\underline{\textbf{Solving $D_i$}}:
We get the partial derivative of Eq.\eqref{eq:lagran} w.r.t. $D_i$, then yields the following closed form:
\begin{equation}\label{eq:Diupdate}
 D_i = (K_1^iU_i - K_3^i + \mu(2X_i - E_i)U_i)\frac{(I + U_i^TU_i)^{-1}}{\mu}
\end{equation}
The major computational burden lies in  $(I + U_i^TU_i)^{-1} \in \mathbb{R}^{c \times c}$, resulting into $\mathcal{O}(c^3)$, which is identical to that for row-updating of $U_i$, hence efficient.\\
\underline{\textbf{Solving $E_i$ }}:
Optimizing Eq.\eqref{eq:lagran} w.r.t. $E_i$  is equivalent to solving the following
\begin{equation}\label{eq:Eiupdate}
\min_{E_i} ||E_i||_1 + \frac{\mu}{2}||E_i - (X_i - D_iU_i^T + \frac{1}{\mu}K_1^{i})||_F^2.
\end{equation}
According to \cite{SJO08},  the following closed form can be obtained
\begin{equation}\label{eq:EiS}
E_i = S_{\frac{1}{\mu}}(X_i - D_iU_i^T + \frac{1}{\mu}K_1^{i}),
\end{equation}
where $S_{\frac{1}{\mu}}(x) = sign(x)max(||x|| - \frac{1}{\mu}, 0)$, $sign(x) = 1$ if $x$ is positive, it is 0 otherwise.\\
\underline{\textbf{Solving $G_i$}}:
Optimizing Eq.\eqref{eq:lagran} w.r.t. $G_i$ is equivalent to the following
\begin{equation}\label{eq:Giup}
\min_{G_i}\Phi(K_2^i, U_i - G_i) + \frac{\mu}{2}||G_i - U_i||_F^2
\end{equation}
Based on that, we enjoy the following closed form
\begin{equation}\label{eq:Giup}
G_i = U_i + \frac{K_2^i}{\mu}
\end{equation}
\underline{\textbf{Solving $W_i$}}:
The problem of optimizing $W_i$ can be converted to the following
\begin{small}
\begin{align}\label{eq:wiup}
& \min_{W_i}\sum_{j,k}\left(\lambda_{2}||U_i(j,\cdot) - U_i(k,\cdot)||_2^2W_i(j,k) + \lambda_{1} W_i(j,k)^2\right)\\ \nonumber
& ~~~~\textmd{s.t.} ~~ \forall j \sum_{k = 1}^{n}W_i(j,k) = 1, 0 \leq W_i(j,k) \leq 1 \\ \nonumber
\end{align}
\end{small}
As the similarity vector for each sample is independent, we only study the $j^{th}$ sample.
\begin{small}
\begin{align}\label{eq:equiw}
& \min\sum_{k}\left(\lambda_{2}||U_i(j,\cdot) - U_i(k,\cdot)||_2^2W_i(j,k) + \lambda_{1} W_i(j,k)^2\right) \\ \nonumber
& ~~~~\textmd{s.t.} ~~  \sum_{k = 1}^{n}W_i(j,k) = 1, 0 \leq W_i(j,k) \leq 1 \\ \nonumber
\end{align}
\end{small}
We convert Eq.\eqref{eq:equiw} to the following
\begin{equation}\label{eq:lastWi}
\min_{\sum_{k = 1}^{n}W_i(j,k) = 1, 0 \leq W_i(j,k) \leq 1}||W_i(j,\cdot) + m_i^j||_2^2,
\end{equation}
where $m_i^j \in \mathbb{R}^{n \times 1}$ is a vector, with its $k^{th}$ entry
$m_i^j(k) = \frac{\lambda_2||U_i(j,\cdot) - U_i(k,\cdot)||_2^2}{4\alpha}$, leading to the following closed form:
\begin{equation}\label{eq:lastclose}
W_i(j,\cdot) = \left(\frac{1 + \sum_{l = 1}^{s}m_i^j(l)}{s}\textmd{\textbf{1}} - m_i^j\right)_+,
\end{equation}
where $(v)_+$ turns the negative entries in $v$ to 0 while with positive entries remained. $s$ denotes the number of data points that have nonzero weight connected to the $j^{th}$ sample. We empirically set $s = 5$ for all views. Once the $\forall j$ $W_i(j,\cdot)$ is obtained, we may update that to be a balanced undirected graph as $\frac{W_i + W_i^T}{2}$.\\
\textbf{\underline{Consensus $W_i(i \in V)$}}: As $W_i$ is solely determined by $U_i$ according to Eq.\eqref{eq:lastclose}, the consensus on $U_i(i \in V)$ in Remark \ref{re:con} naturally leads to the consensus over $W_i(i \in V)$.\\
\textbf{\underline{Multiplier updating}}: The lagrange multipliers $K_1^i$, $K_2^i$ and $K_3^i$ are automatically updated as
\begin{align}\label{eq:lang}
& K_1^i = K_1^i + \mu(X_i - D_iU_i - E_i) \\ \nonumber
& K_2^i = K_2^i + \mu(U_i - G_i)\\ \nonumber
& K_3^i = K_3^i + \mu(D_i - X_iU_i) \nonumber
\end{align}
Besides, $\mu$ is tuned via the adaptive updating rule according to \cite{ALM11}.\\
\textbf{\underline{Algorithm convergence}}:  It is worth nothing that ADM strategy converges to a stationary point yet no guaranteed to be global optimum. Upon that, we define the convergence when $\forall i \in V, ||X_i - X_iU_i - E_i||_F \leq \theta||X||_F$ with $\theta = 10^{-6}$ or maximum iteration number is reached, which is set to be 25 for our method.
\textbf{The optimization process is conducted regarding each variable alternatively within each view, the entire process is terminated until the convergence rule is met for all views. }\\
\textbf{\underline{Multi-view clustering output}}: After the above updating rule is converged, we got the final multi-view clustered representation $U = \sum_{i \in V}U_i \in \mathbb{R}^{n \times c}$; and multi-view optimized local graph structure $W =  \sum_{i \in V}W_i \in \mathbb{R}^{n \times n}$. The normalized graph cut is applied to generate the $c$ clusters as the multi-view spectral clustering output.

We summarize the whole updating process in Algorithm \ref{alg:algorithm1}.
\begin{algorithm}[hbt]
\begin{small}
\KwIn{$X_i (i = 1,\ldots, V), \lambda_1, \lambda_2, \beta$ }
\KwOut{Multi-view spectral clustering result}
\textbf{Initialize}: $U_i[0] (i = 1, \dots,V)$ computation, all entries of $W_i[0], K_1^{i}[0],G_i[0], K_2^{i}[0]$ to be 0, $E_i[0]$ with sparse noise as 20\% entries with uniformly noise over [-5,5], $\mu[0]=10^{-3}$, $k = 0$\\
\For {$i \in V$}{
\textbf{Solve $U_i[k + 1]$}:\\
Sequentially update each row of $U_i[k + 1]$ via Eq.\eqref{eq:Uideri}.\\
Orthonormalizing $U_i[k + 1]$.\\
\textbf{Sequentially update each row of $W_i[k + 1]$} via Eq.\eqref{eq:lastclose}.\\
\textbf{Update $E_i[k + 1]$} via Eq.\eqref{eq:EiS}.\\
\textbf{Update $D_i[k + 1]$} via Eq.\eqref{eq:Diupdate}.\\
\textbf{Update $G_i[k + 1]$} via Eq.\eqref{eq:Giup}.\\
\textbf{Update $K_1^{i}[k + 1]$, $K_2^{i}[k + 1]$, $K_3^{i}[k + 1]$} via Eq.\eqref{eq:lang}.\\
\textbf{Update $\mu$} according to \cite{LinNIPS2011}.\\
\If{\emph{Meet the Convergence Rule}}
{Remove the $i^{th}$ view from the view set as $V = V - i$\\
$U_i[N] = U_i[k + 1]$, s.t. $N$ is any positive integer.\\
}
\Else{$k = k + 1$}
}

$U = \sum_{i \in V}U_i[k + 1], W = \sum_{i \in V}W_i[k + 1]$ ($i=1,\ldots,V$)
\end{small}
\textbf{Return} Multi-view spectral clustering results based on $U$ and $W$ via normalized graph cut.
\caption{\small Solving Eq.\eqref{eq:finalobj} for multi-view spectral clustering.}\label{alg:algorithm1}\small
\end{algorithm}

\section{Experimental Validation}\label{sec:experiment}
The following multi-view data sets and their view-specific features are selected according to \cite{RMVSC,ijcai16}.
\begin{itemize}
\item \underline{UCI handwritten Digit set}\footnote{http://archive.ics.uci.edu/ml/datasets/Multiple+Features}: It consists of features of hand-written digits (0-9). The dataset is described by 6 features and contains 2000 samples with 200 in each category. Analogous to \cite{ALM11}, we choose 76 Fourier coefficients  (FC) of the character shapes and the 216 profile correlations (PC) as two views.

\begin{table}
\caption{Multi-view consensus ratio metric as per Eq.\eqref{eq:metriclast} between our method and \textbf{LRRGL} over three data sets. Smaller value means similar magnitude. }
\begin{center}
\begin{tabular}{cccc}
\hline\hline
\textsf{Method} & UCI digits & AwA & NUS \\
\hline
\textbf{LRRGL} & 17.39 & 25.78 & 34.21\\
\textbf{Ours} & \textbf{1.15} & \textbf{1.18}& \textbf{1.21}\\
\hline
\end{tabular}
\end{center}
\label{table:ratiometric}
\end{table}

\item \underline{Animal with Attribute} (AwA)\footnote{http://attributes.kyb.tuebingen.mpg.de}: It consists of 50 kinds of animals described by 6 features (views): Color histogram ( CQ, 2688-dim),  local self-similarity (LSS, 2000-dim),  pyramid HOG (PHOG, 252-dim), SIFT (2000-dim), Color SIFT (RGSIFT, 2000-dim), and SURF (2000-dim). We randomly sample 80 images for each category and get 4000 images in total.

\item \underline{NUS-WIDE-Object (NUS)} \cite{NUS-Wide}: The data set consists of 30000 images from 31 categories. We construct 5 views: 65-dimensional color histogram (CH), 226-dimensional color moments (CM), 145-dimensional color correlation (CORR), 74-dimensional edge estimation (EDH), and 129-dimensional wavelet texture (WT).
\end{itemize}
The following typical multi-view baselines are compared for spectral clustering, covering \emph{\underline{Early fusion}}, \emph{\underline{Late fusion}}, \emph{\underline{CCA}}, \emph{\underline{Co-training}} strategy and \emph{\underline{LRR models}} as reviewed in Section ~\ref{sec:related}. All the parameters are tuned to their best performance.
\begin{itemize}
\item \textbf{MFMSC}: Concatenating multi-view features to perform spectral clustering.

\item Multi-view affinity aggregation for multi-view spectral clustering (\textbf{MAASC}) \cite{CVPR12}.

\item Canonical Correlation Analysis (CCA) based multi-view spectral clustering (\textbf{CCAMSC}) \cite{ICML09} by learning a common subspace for multi-view data, then perform spectral clustering.

\item \textbf{Co-training} \cite{icml11}: Learning multi-view Laplacian eigenspace via a co-training fashion over each individual one.

\item Robust Low-Rank Representation Method (\textbf{RLRR}) \cite{RMVSC}, as formulated in Eq.\eqref{eq:motivate}.

\item Low-Rank Representation with Multi-Graph Learning (\textbf{LRRGL}) \cite{ijcai16}.
\end{itemize}

\begin{table}[t]
\caption{Averaged clustering results in terms of \textsf{ACC} on three benchmark data sets.}
\begin{center}
\begin{tabular}{cccc}
\hline\hline
\textsf{ACC} (\textbf{\%}) & UCI digits & AwA & NUS \\
\hline
\textbf{MFMSC} & 43.81 & 17.13& 22.81\\
\textbf{MAASC} & 51.74 & 19.44& 25.13\\
\textbf{CCAMSC} & 73.24 & 24.04& 27.56\\
\textbf{Co-training} & 79.22 & 29.06 & 34.25\\
\textbf{RLRR} & 83.67 & 31.49 & 35.27\\
\textbf{LRRGL} & 86.39 & 37.22 & 41.02\\
\hline
\textbf{Ours} & \textbf{92.22} & \textbf{44.55} & \textbf{45.78} \\
\hline
\end{tabular}
\end{center}
\label{table:acc}
\end{table}

\begin{table}
\caption{Averaged clustering results in terms of \textsf{NMI} on three benchmark data sets.}
\begin{center}
\begin{tabular}{cccc}
\hline\hline
\textsf{NMI} (\textbf{\%}) & UCI digits & AwA & NUS \\
\hline
\textbf{MFMSC} & 41.57 & 11.48 & 12.21\\
\textbf{MAASC} & 47.85 & 12.93& 11.86\\
\textbf{CCAMSC} & 56.51 & 15.62& 14.56\\
\textbf{Co-training} & 62.07 & 18.05 & 18.10\\
\textbf{RLRR} & 81.20 & 25.57 & 18.29\\
\textbf{LRRGL} & 85.45 & 31.74 & 20.61\\
\hline
\textbf{Ours} & \textbf{89.61} & \textbf{36.67} & \textbf{26.42} \\
\hline
\end{tabular}
\end{center}
\label{table:nmi}
\end{table}

\begin{figure}[t]
\begin{center}
\begin{tabular}{cc}
\includegraphics[width=3.7cm,height=3.7cm]{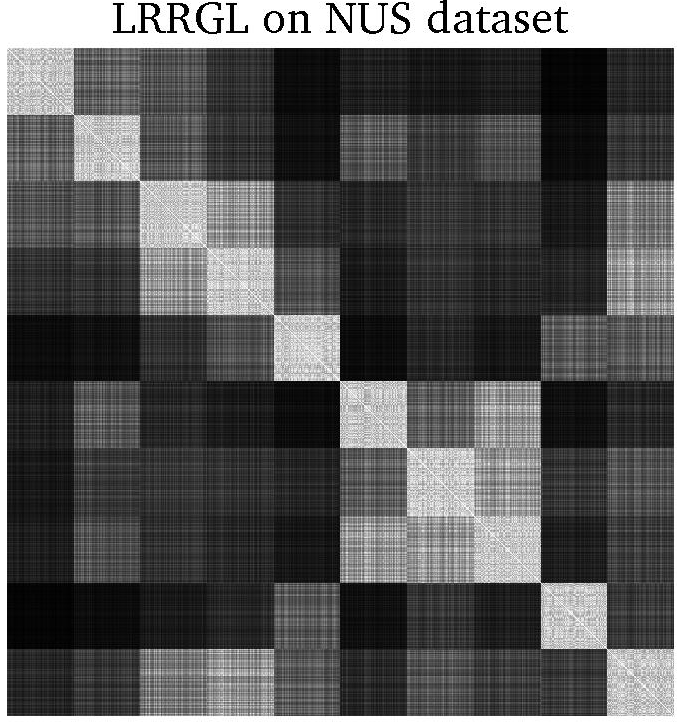}&
\includegraphics[width=3.7cm,height=3.7cm]{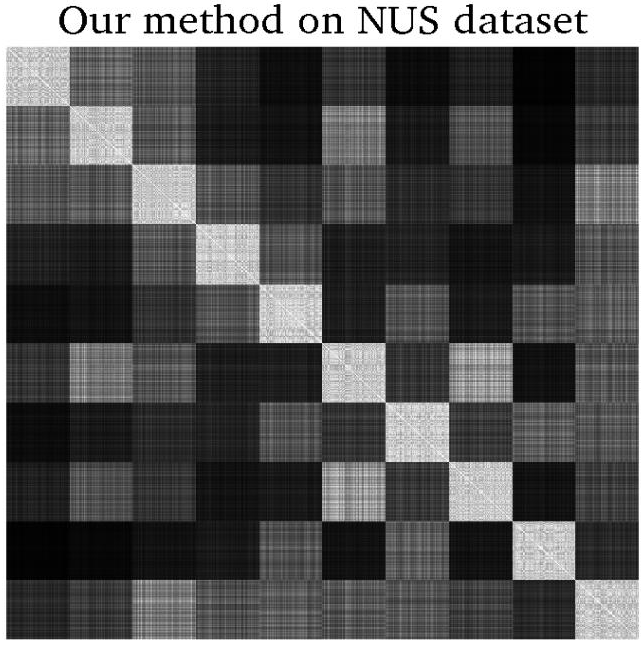}\\
(a)&(b)
\end{tabular}
\end{center}
\caption{The visualized multi-view data similarity consensus result between ours and \textbf{LRRGL} over NUS data set. We randomly select 10 classes, where 80 samples are randomly selected for each of them. The 10 diagonal block represents the data samples within the 10 ideal clusters, where the whiter the color is, the ideally affinity value will be. Meanwhile, for non-diagonal blocks,
the more black the color is, the smaller affinity will be to reveal the different clusters.}\label{fig:affi}
\end{figure}

\begin{figure}
\begin{center}
\begin{tabular}{cc}
\includegraphics[width=4.1cm,height=4.1cm]{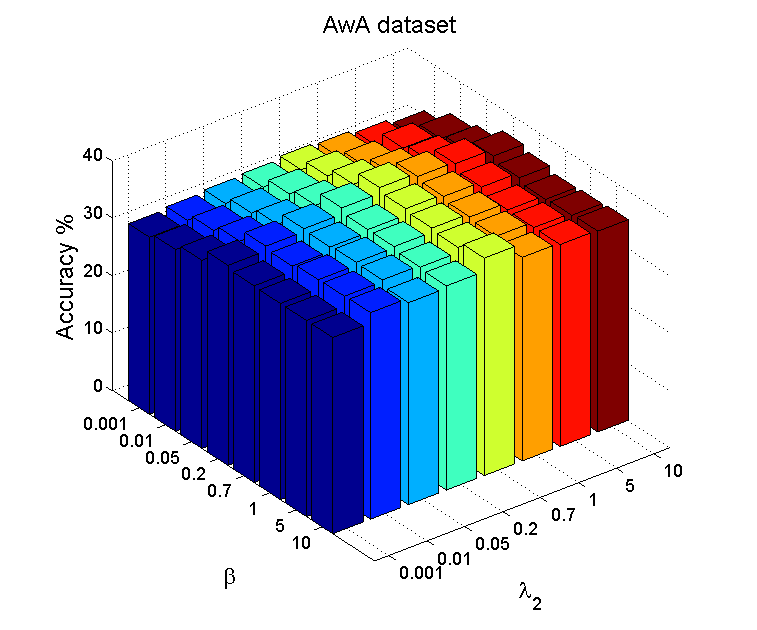}&
\includegraphics[width=4.1cm,height=4.1cm]{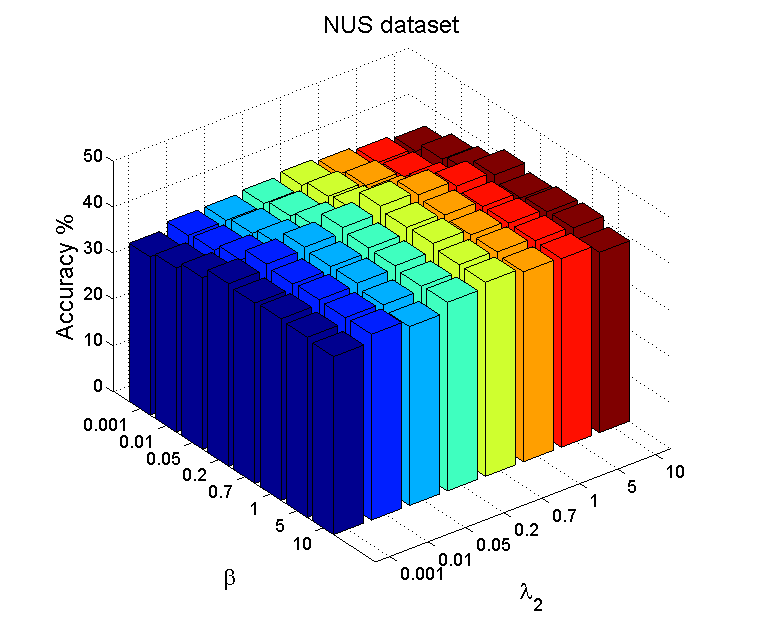}\\
(a)&(b)
\end{tabular}
\end{center}
\caption{Parameters study of $\lambda_2$ and $\beta$ for our method over NUS and AwA data sets evaluated with clustering accuracy. }\label{fig:accuparatest}
\end{figure}

Clustering accuracy (\textsf{ACC}) and normalized mutual information (\textsf{NMI}). Pleaser refer to \cite{SKMEA,Chen-TPAMI11} for their detailed descriptions. To demonstrate the robustness superiority over non-LRR methods,  following \cite{ijcai16}, we set the feature corruption noise for each view is with sparse noise as 20\% entries with uniformly noise over [-5,5] for \textbf{RLRR}, \textbf{LRRGL} and our method, with $\lambda_1 = 0.8$ in Eq.\eqref{eq:finalobj} for our method. All experiments are repeated 10 times,  the average clustering results are shown in Tables \ref{table:acc} and \ref{table:nmi}, where our method outperforms the others, especially better than \textbf{RLRR} and \textbf{LRRGL}, due to its strengthes of
\begin{itemize}
\item encoding more flexible latent cluster structures, along with the more ideal optimized local graph structure based on such latent clustered representation.
\item The superior multi-view consensus in terms of both latent clustered representation and optimized local graph structure for all views.
\end{itemize}
To penetrate the first finding, we illustrate the visualized consensus multi-view affinity matrix over NUS data set between our method and \textbf{LRRGL} in Fig.~\ref{fig:affi}, which validates the advantages of our clustered orthogonal representation over low-rank similarity yielded by \textbf{LRRGL}.\\
\textbf{\underline{Parameter Study:}} We further study the parameter $\lambda_2$ (clustered orthogonal representations and optimized local graph structure) and $\beta$ (multi-view consensus term) in Eq.\eqref{eq:finalobj}, and against the clustering accuracy over AwA and NUS data sets; we varied one parameter while fixed the others, and the results are illustrated in Fig.~\ref{fig:accuparatest}, where increasing either of them can improve the clustering accuracy until meet the optimal pair-wise values, followed by a slight performance decreasing. To balance Figs. ~\ref{fig:accuparatest}(a) and (b), we finalize $\lambda_2 = 0.7$ and $\beta = 0.25$ in Eq.\eqref{eq:finalobj}.

\section{Conclusion}
In this paper, we revisit the classical Low-Rank Representation (LRR) for multi-view spectral clustering, by viewing LRR as essentially a latent clustered orthogonal projection winged with its optimized local graph structure.  Following this, we propose to
simultaneously learn clustered orthogonal projection and optimized local graph structure for each view, while enjoy the same magnitude over them both for all views, leading to a superior multi-view spectral clustering consensus. Extensive experiments validate its strength.

{
\bibliographystyle{named}
\bibliography{ijcai17}
}
\end{document}